\pdfoutput=1

\documentclass[11pt]{article}

\usepackage{emnlp2021}

\usepackage{times}
\usepackage{latexsym}
\usepackage{graphicx}
\usepackage[nolist]{acronym}                
\usepackage{hyperref}                       
\usepackage{xurl}                           
\usepackage{booktabs}                       
\usepackage{amsfonts}                       
\usepackage{nicefrac}                       
\usepackage{enumitem}                       
\usepackage{pgfplots,pgfplotstable}                       
\pgfplotsset{compat=1.17} 

\usepackage[T1]{fontenc}

\usepackage[utf8]{inputenc}

\usepackage{microtype}

\begin{acronym}[UMLX]
    \acro{FSCS}{Federal Supreme Court of Switzerland}
    \acro{ECtHR}{European Court of Human Rights}
    \acro{US}{United States}
    \acro{SCOTUS}{Supreme Court of the United States}
    \acro{SPC}{Supreme People's Court of China}

    \acro{RegExp}{Regular Expression}
    
    \acro{LSTM}{Long Short-Term Memory}
    \acro{BiLSTM}{Bidirectional Long Short-Term Memory}
    
    \acro{NLP}{Natural Language Processing}
    \acro{ML}{Machine Learning}
    \acro{DAG}{Directed Acyclic Graph}
    \acro{LJP}{Legal Judgment Prediction}
    \acro{BoW}{Bag-of-Words}
    \acro{TF-IDF}{Term Frequency - Inverse Document Frequency}
    \acro{SVM}{Support Vector Machine}
    \acro{LR}{Logistic Regression}

\end{acronym}

\title{Swiss-Judgment-Prediction: A Multilingual Legal \\Judgment Prediction Benchmark}

\author{
  Joel~Niklaus\\
  Research Center for Digital Sustainability\\
  Institute of Computer Science\\
  University of Bern\\
  Bern, Switzerland \\
  \texttt{joel.niklaus@inf.unibe.ch} \\
  \And
  Ilias~Chalkidis \\
  Coastal NLP Group\\
  Department of Computer Science\\
  University of Copenhagen\\
  Copenhagen, Denmark \\
  \texttt{ilias.chalkidis@di.ku.dk} \\
  \AND
  Matthias~Stürmer \\
  Research Center for Digital Sustainability\\
  Institute of Computer Science\\
  University of Bern\\
  Bern, Switzerland \\
  \texttt{matthias.stuermer@inf.unibe.ch} \\
}

\begin{document}
\maketitle

\begin{abstract}

    In many jurisdictions, the excessive workload of courts leads to high delays. Suitable predictive AI models can assist legal professionals in their work, and thus enhance and speed up the process.
    So far, \ac{LJP} datasets have been released in English, French, and Chinese. We publicly release a multilingual (German, French, and Italian), diachronic (2000-2020) corpus of 85K cases from the \ac{FSCS}. We evaluate state-of-the-art BERT-based methods including two variants of BERT that overcome the BERT input (text) length limitation (up to 512 tokens). Hierarchical BERT has the best performance (approx. 68-70\% Macro-F1-Score in German and French). Furthermore, we study how several factors (canton of origin, year of publication, text length, legal area) affect performance. We release both the benchmark dataset and our code to accelerate future research and ensure reproducibility.

\end{abstract}

\section{Introduction}
\label{sec:introduction}

Frequently, legal information is available in textual form (e.g. court decisions, laws, legal articles or commentaries, contracts). With the abundance of legal texts comes the possibility of applying \ac{NLP} techniques to tackle challenging tasks \cite{Chalkidis2018DeepLI,zhong-etal-2020-nlp,chalkidis-etal-2021-lexglue}.
In this work, we study the task of Legal Judgment Prediction (\ac{LJP}) where the goal is to predict the outcome (verdict) of a decision given its facts \cite{aletras_predicting_2016, sulea_predicting_2017, luo_learning_2017, zhong_legal_2018, hu_few-shot_2018, chalkidis_neural_2019}. 
Many relevant applications and tasks, such as  court opinion generation \cite{ye_interpretable_2018} and analysis \cite{wang_historical_2012} have been also studied, while there is also work aiming to interpret (explain) the decisions of particular courts \cite{ye_interpretable_2018,chalkidis-et-al-2021-ecthr}.

Models developed for \ac{LJP} and relevant supportive tasks may assist both lawyers, e.g., help them prepare their arguments by identifying their strengths and weaknesses, and judges and clerks, e.g., review or prioritize cases, thus speeding up judicial processes and improving their quality. Especially in areas with many pending cases such as Indian\footnote{\url{https://tinyurl.com/mjy2uf9a}} and Brazilian\footnote{\url{https://tinyurl.com/2uttucmn}} jurisdictions or US immigration cases\footnote{\url{https://tinyurl.com/4ybhhff8}} the deployment of such models may drastically shorten the backlog. Such models can also help legal scholars to study case law \cite{Katz2012} and help sociologists and research ethicists to expose irresponsible use of AI in the justice system \cite{angwin2016, dressel2018}.
So far, \ac{LJP} datasets have been released for English \cite{katz_general_2017, Medvedeva2018, chalkidis_neural_2019},  French \cite{sulea_predicting_2017} and Chinese \cite{ xiao_cail2018_2018, long_automatic_2019}.

We introduce a new multilingual, diachronic \ac{LJP} dataset of \ac{FSCS} cases, which spans 21 years (from 2000 to 2020) containing over 85K (50K German, 31K French and 4K Italian) cases.
To the best of our knowledge, it is the only publicly available multilingual \ac{LJP} dataset to date.
Additionally, it is annotated with publication years, legal areas and cantons of origin; thus it can be used also as test-bed for fairness and robustness in the critical application of \ac{NLP} to law \cite{Wang2021EqualityBT}. 

\citet{rogers_changing_2021} argues that the \ac{NLP} community is investing many more resources in the development of models rather than data. As a result, there are not enough challenging, high-quality and well curated benchmarks available. \citeauthor{rogers_changing_2021} assumes that the main reason for this imbalance is that the ''data work`` is considered less prestigious and top conferences are more likely to reject resource (dataset) papers. With our work (and the associated code and data) we hope to make a valuable contribution to the legal \ac{NLP} field, where there are not many ready-to-use benchmarks available.

\subsection*{Contributions}
The contributions of this paper are threefold:
\begin{itemize}[leftmargin=8pt]
    \item We publicly release a large, high quality, curated, multilingual, diachronic dataset of 85K Swiss  Federal  Supreme  Court (\ac{FSCS}) cases annotated with the respective binarized judgment outcome (\emph{approval}/\emph{dismissal}), posing a challenging text classification task. We also provide additional metadata, i.e., the publication year, the legal area and the canton of origin per case, to promote robustness and fairness studies on the critical area of legal \ac{NLP} \cite{Wang2021EqualityBT}.
    
    \item We provide experimental results with strong baselines representing the current state-of-the-art in \ac{NLP}. 
    Since the average length of the facts (850 tokens in the French part) is longer than the 512 tokens limit by BERT \cite{devlin_bert_2019}, special methods are needed to cope with that. We show results comparing standard BERT models (up to 512 tokens) with two variants (hierarchical and prolonged BERT) that use up to 2048 tokens.
    
    \item We analyze the results of the German dataset in terms of diachronicity (publication year), legal area and input (text) length and the French dataset by canton of origin. We find that performance deteriorates as cases are getting more complex (longer facts), while also performance varies across legal areas. There is no sign of performance fluctuation across years.
    
\end{itemize}

\section{Related Work}
\label{sec:related_work}

\subsection*{\ac{ECtHR}}
\citet{aletras_predicting_2016} introduced a dataset of 584 \ac{ECtHR} cases concerning the violation or not of three articles of the European Convention of Human Rights (ECHR). They used a \ac{SVM} \cite{cortes1995support} with \ac{BoW} (n-grams) and topical features  on a simplified binarized \ac{LJP}. 
In contrast to our work, they evaluated with random 10-fold cross-validation instead of the more realistic temporal split based on the date \cite{sogaard2021}. 
\citet{Medvedeva2018} extended the \ac{ECtHR} dataset to include 9 instead of 3 Articles resulting in a total of approx. 11.5K cases. They also experimented with an \ac{SVM} operating on n-grams on the \ac{LJP} task.
\citet{chalkidis_neural_2019} experimented on a similarly sized dataset using neural methods. On the binary \ac{LJP} task, they improve the state-of-the-art
using a hierarchical version of BERT. Additionally, they experimented with a multi-label \ac{LJP} task predicting for each of the 66 ECHR Articles whether it is violated or not.\vspace{2mm} 

\subsection*{\ac{SCOTUS}}
\citet{katz_general_2017} experimented on \ac{LJP} with 28K cases from the \ac{SCOTUS} spanning almost two centuries. They trained a Random Forest \cite{breiman2001random} classifier using extensive feature engineering with many non textual features.
\citet{kaufman_kraft_sen_2019} improved results using an ADABoost \cite{FREUND1997119} classifier, while also incorporating more textual information (i.e., statements made by the court judges during oral arguments).\vspace{2mm}

\subsection*{French Supreme Court (Court of Cassation)}
\citet{sulea_predicting_2017} studied the \ac{LJP} task on a dataset of approx. 127K French Supreme Court cases. They experimented on a 6-class and a 8-class setting using an \ac{SVM} with \ac{BoW} features. They reported very high scores, which they claim are justified by the high predictability of the French Supreme Court. Although they used as input the entire case description and not only the facts, thus there is a strong possibility of label information leak. They also used 10-fold stratified cross-validation selecting the test part at random. \vspace{2mm}

\subsection*{German Courts}
\citet{urchs_design_2021} present a corpus of over 32K German court decisions from 131 Bavarian courts. The corpus is annotated with rich metadata including, among others, facts and judgment outcome needed for the \ac{LJP} task. They present sample experiments predicting the type of the decision (judgment, resolution or other) and detecting conclusion, definition and subsumption in a subset of 200 randomly chosen and manually annotated decisions. They used traditional \ac{ML} methods such as \ac{LR} on unigrams (\ac{BoW} features) and \ac{SVM} on \ac{TF-IDF} features.

\subsection*{\ac{SPC}}
\citet{luo_learning_2017} experimented with the Hierarchical Attention Network \cite{yang-etal-2016-hierarchical} on Chinese criminal cases. They trained a model jointly on charge prediction, a form of \ac{LJP}, and the relevant criminal law article extraction task using the relevant articles as support for the charge prediction.
\citet{xiao_cail2018_2018} introduced a large-scale \ac{LJP} dataset of more than 2.6M Chinese criminal cases from the \ac{SPC}. Their dataset is annotated with extensive metadata such as  applicable law articles, charges, and prison terms.
\citet{zhong_legal_2018} viewed the dependencies between the different subtasks of \ac{LJP} as a \ac{DAG} and apply a topological multitask learning framework. They work on three different datasets each containing Chinese criminal cases.
\citet{long_automatic_2019} studied the \ac{LJP} task on 100K Chinese divorce proceedings considering three types of information as input: applicable law articles, fact description, and plaintiffs' pleas. 
\citet{li_mann_2019} use a multichannel attentive neural network on four datasets containing Chinese criminal cases. They considered all three subtasks of the Chinese \ac{LJP} datasets: charges, law articles and prison term.
\citet{yang_recurrent_2019} apply a recurrent attention network on three Chinese \ac{LJP} datasets. \vspace{2mm}

\section{Data Description}
\label{sec:data_description}

\subsection{Dataset Construction}

The decisions were downloaded from the platform entscheidsuche.ch and have been pre-processed by the means of HTML parsers and \acp{RegExp}. The dataset contains more than 85K decisions from the \ac{FSCS} written in three languages (50K German, 31K French, 4K Italian) from the years 2000 to 2020.\footnote{The dataset is not parallel, all cases are unique and decision are written only in a single language.} The \ac{FSCS} is the last level of appeal in Switzerland and hears only the most controversial cases which could not have been sufficiently well solved by (up to two) lower courts. In their decisions, they often focus only on small parts of previous decision, where they discuss possible wrong reasoning by the lower court. This makes these cases particularly challenging.

In order to fight the reproducibility crisis \cite{britz_ai_2020}, we release the Swiss-Judgment-Prediction dataset on Zenodo\footnote{\url{https://zenodo.org/record/5529712}} and on Hugging Face\footnote{\url{https://huggingface.co/datasets/swiss_judgment_prediction}}, while also open-sourcing the complete code used for constructing the dataset\footnote{\url{https://github.com/JoelNiklaus/SwissCourtRulingCorpus}} as well as for running the experiments\footnote{\url{https://github.com/JoelNiklaus/SwissJudgementPrediction}} on GitHub.

\subsection{Structure of Court Decisions}
A typical Swiss court decision is made up of the following four main sections: \emph{rubrum}, \emph{facts}, \emph{considerations} and \emph{rulings}.\footnote{See examples in Figures~\ref{fig:dismissed_decision} and \ref{fig:approved_decision} of Appendix \ref{sec:examples}.} 
The \emph{rubrum} (introduction) contains the date and chamber, mentions the involved judge(s) and parties and finally states the topic of the decision.
The \emph{facts} describe what happened in the case and form the basis for the considerations of the court. The higher the level of appeal, the more general and summarized the facts.
The \emph{considerations} reflect the formal legal reasoning which form the basis for the final ruling. Here the court cites laws and other influential rulings. 
The \emph{rulings}, constituting the final section, are an enumeration of the binding decisions made by the court. This section is normally rather short and summarizes the considerations.

\begin{table*}[!ht]
\resizebox{\textwidth}{!}{%
\begin{tabular}{@{}lrrrcrrrcrrr@{}}
\toprule
Split & \multicolumn{3}{c}{de} & \phantom{a}& \multicolumn{3}{c}{fr} &
\phantom{a} & \multicolumn{3}{c}{it}\\
\cmidrule{2-4} \cmidrule{6-8} \cmidrule{10-12}
& approval & dismissal & total && approval & dismissal & total && approval & dismissal & total\\ 
\midrule
train   & 8369 (24\%)   & 27003 (76\%) & 35452 && 5197 (25\%)   & 15982 (75\%) & 21179 && 625 (20\%) & 2447 (80\%)  & 3072\\ 
val     & 959 (20\%)    & 3746 (80\%)  & 4705  && 649 (21\%)    & 2446 (79\%)  & 3095  && 65 (16\%)  & 343 (84\%)   & 408\\
test    & 1915 (20\%)   & 7810 (80\%)  & 9725  && 1264 (19\%)   & 5556 (81\%)  & 6820  && 152 (19\%) & 660 (81\%)   & 812\\
all     & 11243 (23\%)  & 38639 (77\%) & 49882  && 7110 (23\%)   & 23984 (77\%) & 31094 && 842 (20\%) & 3450 (80\%)  & 4292\\
\bottomrule
\end{tabular}
}
\vspace{-2mm}
\caption{
The number of cases per label (\emph{approval}, \emph{dismissal}) in each language subset. 
}
\label{tab:label_imbalance}
\vspace{-5mm}
\end{table*}

\subsubsection{Use of Facts instead of Considerations}
\label{sec:facts_vs_considerations}

We deliberately did not consider the considerations as input to the model, unlike \citet{aletras_predicting_2016} for the following reasons.
The facts are the section which is most similar to a general description of the case, which may be more widely available, while being less biased.\footnote{Note however, that the facts are drafted together with the considerations and are often formulated in a way to support the reasoning in the considerations.}
Additionally, the facts do not change that much from one to the next level of appeal (apart from being more concise and summarized in the higher levels of appeal).
According to estimations from several court clerks we consulted, the facts take approximately 10\% of the time for drafting a decision while the considerations take 85\% and the outcome 5\% (45\%, 50\% and 5\% in penal law  respectively).
So, most of the work being done by the judges and clerks results in the legal considerations. Therefore, we would expect the model to perform better if it had access to the considerations. But on the other hand, the value of the model would be far smaller, since most of the work is already done, once the considerations are written. Thus, to create a more realistic and challenging scenario, we consider only the facts as input for the predictive models.

\subsection{The Binarized \ac{LJP} Task - Verdict Labeling Simplification}
\label{sec:labels}

The cases have been originally labeled with 6 labels: \emph{approval}, \emph{partial approval}, \emph{dismissal}, \emph{partial dismissal}, \emph{inadmissible} and \emph{write off}.
The first four are judged on the basis of the facts (merits) and the last two for formal reasons.
A case is considered \emph{inadmissible}, if there are formal deficiencies with the appeal or if the court is not responsible to rule the case.
A court rules \emph{write off} if the case has become redundant so there is no reason for the proceeding anymore.
This can be for several reasons, such as an out-of-court settlement or procedural association (two proceedings are unified).
\emph{Approval} and \emph{partial approval} mean that the request is deemed valid or partially valid respectively.
\emph{Dismissal} and \emph{partial dismissal} mean that the request is denied or partially denied respectively.
A \emph{partial} decision is usually ruled in parallel with a decision of the opposite kind or with \emph{inadmissible}. 

In practice, court decisions may have multiple requests (questions), where each can be judged individually. 
Since the structure of the outcomes in the decisions is non-standard, parsing them automatically is very challenging. 
Therefore, we decided to focus on the main request only and discard all side (secondary) requests.
Even the main request sometimes contains multiple judgments referring to different parts of the main request, with some more important than others (it is very hard to automatically detect their criticality). 
So, to simplify the task and make it more concise, we transform the document labeling from a list of partial judgments into a single judgment, as follows:
\begin{enumerate}[leftmargin=8pt, noitemsep]
    \item We excluded all cases that have been ruled with both an approval and a dismissal in the main request, since that could be rather confusing.
    \item We excluded cases ruled with \emph{write off} outcomes since these cases are rejected for formal reasons that are not written (described) in the facts. Therefore, a model has no chance of inferring it correctly. We also excluded cases with \emph{inadmissible} outcomes for similar reasons. 
    \item Since \emph{partial} approvals/dismissals are very hard to distinguish from \emph{full} approvals/dismissals respectively, we converted all the partial ones to full ones. Thus, the final labeling includes two possible outcomes, approvals and dismissals (i.e., the court ``leans'' positive or negative to the request).
\end{enumerate}

\begin{figure*}[t]
\centering
\includegraphics[width=\textwidth, height=5cm]{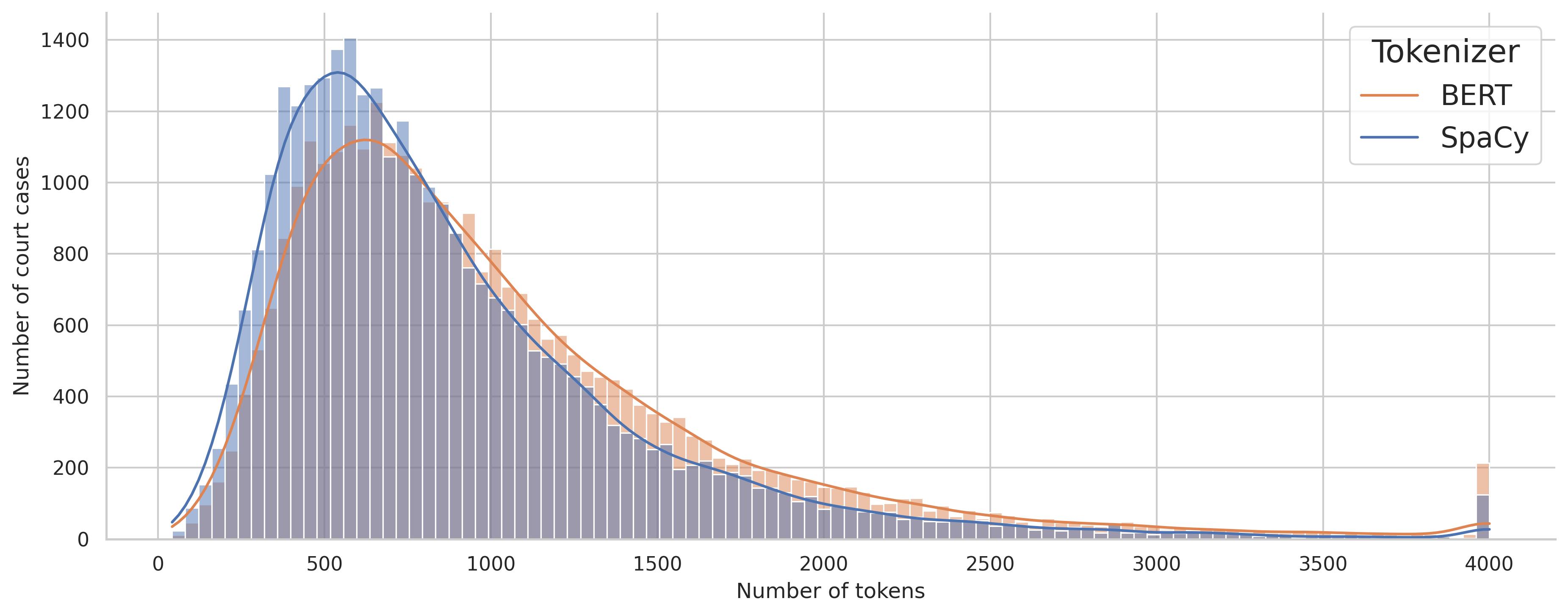}
\caption{The distribution of the document (the facts of a case) length for French decisions.
The blue histogram shows the document (case) length distribution in regular words (using the SpaCy tokenizer \protect\cite{honnibal_spacy_2020}). It is useful for a human estimation of the length and for methods building upon word embeddings \protect\cite{mikolov_efficient_2013,pennington_glove_2014}. The orange histogram shows the distribution in sub-word units (generated by the SentencePiece tokenizer \protect\cite{kudo_sentencepiece_2018} used in BERT). It is useful e.g. for estimating the maximum sequence length of a BERT-like model.
Decisions with length over 4000 tokens have been grouped in the last bin.}
\label{fig:input_length_dist_fr}
\vspace{-4mm}
\end{figure*}

By implementing these simplifications, we made the dataset more feasible (solvable) and semantically coherent targeting the core ruling process (see Section \ref{sec:experiments}). Table \ref{tab:num_decisions} shows the numbers of decisions after each processing step. Note that we reduced the dataset with these preprocessing steps significantly (from over 141K to close to 85K decisions) to achieve higher quality. We also made the task structurally simpler by converting it from a multi-label to a binary classification task.\footnote{Although, we look forward to recover at least part of the complexity in the future, if we have the appropriate resources to manually extract per-request judgments, introducing a new multi-task (multi-question) \ac{LJP} dataset.}

The dataset is highly imbalanced containing more than $\frac{3}{4}$ dismissed cases (see Table \ref{tab:label_imbalance} for details). The label skewness makes the classification task quite hard and beating dummy baselines, e.g., predicting always the majority class, on micro-averaged measures (e.g., Micro-F1) is challenging. In our opinion, macro-averaged measures (e.g., Macro-F1) are more suitable in this setting, since they consider both outcomes (classes); they can also better discriminate better methods. In other words, they favor models that can actually learn the task (discriminate the two classes) and they do not always predict the majority class, i.e., \emph{dismissal}, regardless of the facts.

\begin{table}[!th]
\resizebox{\columnwidth}{!}{%
\begin{tabular}{@{}lrrrrr@{}}
\toprule
Language & Total & 2000-2020 & Rulings    & Judgments & Binarized\\ 
\midrule
de       & 96337  & 95449      & 95273      & 84083     & 49882\\
fr       & 52278  & 51748      & 49132      & 49083     & 31094\\
it       & 8784   & 8643       & 8457       & 8441      & 4292\\
all      & 157399 & 155840     & 152862     & 141607    & 85268\\
\bottomrule
\end{tabular}
}
\caption{
\emph{Rulings} is the number of cases where rulings could be extracted.
\emph{Judgments} is the number of cases where we could extract any judgment types described in Section \ref{sec:labels}.
\emph{Binarized} is the number of cases considered in the final dataset after removing decisions containing labels other than \emph{approval} or \emph{dismissal}.
}
\label{tab:num_decisions}
\vspace{-4mm}
\end{table}

\subsection{Case Distribution}

This Section presents statistics about the distribution of cases according to different metadata like input (text) length, legal area and origin cantons.

\subsubsection{The Curse of Long Documents}
\label{sec:long_documents}

Figure \ref{fig:input_length_dist_fr} shows the distribution of the document (facts of the case) length of French cases.\footnote{See Figures \ref{fig:input_length_dist_de} and \ref{fig:input_length_dist_it} in Appendix \ref{sec:input_length_dist} for the German and Italian cases, respectively.} We see that there are very few decisions with more than 2K tokens in German (very similar for Italian). The French decisions are more evenly distributed, including a large portion of decisions with more than 4K tokens. For all languages, there is a considerable portion of decisions (50\%+) containing more than 512 sub-word units (BERTs maximum sequence length) posing a fundamental challenge for standard BERT models.

\subsubsection{Legal Areas}

Table \ref{tab:legal_area_distribution} presents the distribution of legal areas across languages. The legal areas are derived from the chambers where the decisions were heard. The website of the \ac{FSCS}\footnote{\url{https://tinyurl.com/52a4x8yz} (in German)} describes in detail what kinds of cases the different chambers hear.

\begin{table}[!ht]
\resizebox{\columnwidth}{!}{%
\begin{tabular}{@{}lrrr@{}}
\toprule
Legal Area  & de            & fr            & it\\
\midrule
public law     & 12182 (24\%)  & 8514  (27\%)  & 1583 (37\%)\\
penal law       & 10942 (22\%)  & 8039  (26\%)  & 692 (16\%)\\
social law      & 10742 (22\%)  & 4048  (13\%)  & 673 (16\%)\\
civil law       & 8208  (16\%)  & 7348  (24\%)  & 763 (18\%)\\
insurance law   & 7625  (15\%)  & 2950  \phantom{a}(9\%)   & 561 (13\%)\\
other       & 183  (0.4\%)   & 195  (0.6\%)   & 20  (0.5\%)\\
\bottomrule
\end{tabular}
}
\caption{
The distribution of legal areas in each language subset.
}
\label{tab:legal_area_distribution}
\vspace{-4mm}
\end{table}

\subsubsection{Origin Cantons}

To study robustness and fairness in terms of geographical (regional) groups, we extracted the canton of origin from the decisions. As we observe in Table \ref{tab:origin_canton_distribution}, most of the cantons (e.g., Zürich, Ticino) are monolingual and the distribution of the cases across cantons is very skewed with 1-2 cantons per language covering a large portion of the total cases.

\begin{table}[!ht]
\resizebox{\columnwidth}{!}{%
\begin{tabular}{@{}lrrr@{}}
\toprule
Canton of Origin            & de            & fr            & it\\
\midrule
Zürich (ZH)                 & \textbf{12749 (25\%)}  & -             & - \\
Berne (BE)                  & 4705 (9\%)  & \underline{469 (2\%)}              & - \\
Lucerne (LU)                & 3124 (6\%)  & -             & - \\
Uri (UR)                    & \underline{248 (0.5\%)}  & -             & - \\
Schwyz (SZ)                 & \underline{1408 (3\%)}  & -             & - \\
Obwalden (OW)               & \underline{190 (0.4\%)}  & -             & - \\
Nidwalden (NW)              & \underline{364 (0.7\%)}  & -             & - \\
Glarus (GL)                 & \underline{363 (0.7\%)}  & -             & - \\
Zug (ZG)                    & \underline{1321 (3\%)}  & -             & - \\
Fribourg (FR)               & \underline{487 (1\%)}  & 1826 (6\%)       & - \\
Soleure (SO)                & \underline{2022 (4\%)}    & -            & - \\
Basel-City (BS)             & \underline{1651 (3\%)}  & -             & - \\
Basel-Country (BL)          & \underline{1578 (3\%)}  & -             & - \\
Schaffhausen (SH)           & \underline{591 (1\%)}  & -             & - \\
Appenzell Outer-Rhodes (AR) & \underline{73 (0.2\%)}  & -             & - \\
Appenzell Inner-Rhodes (AI) & \underline{103 (0.2\%)}  & -             & - \\
St. Gall (SG)               & 3188 (6\%)  & -             & - \\
Grisons (GR)                & \underline{1300 (3\%)} & -             & \underline{85 (2\%)} \\
Argovia (AG)                & 5494 (11\%)  & -             & - \\
Thurgovia (TG)              & \underline{2066 (4\%)}  & -             & - \\
Ticino (TI)                 & -             & -             & \textbf{3302 (77\%)} \\
Vaud (VD)                   & -             & \textbf{8926 (29\%)} & - \\
Valais (VS)                 & \underline{502 (1\%)} & 2095 (7\%)   & - \\
Neuchâtel (NE)              & -             & 1732 (6\%)   & - \\
Genève (GE)                 & -             & \textbf{9320 (30\%)} & - \\
Jura (JU)                   & -             & \underline{630 (2\%)}   & - \\
Swiss Confederation (CH)    & \underline{1854 (4\%)}  & \underline{348 (1\%)}       & \underline{83 (2\%)} \\
uncategorized               & 4488 (9\%)  & 5742 (18\%)       & 818 (19\%) \\
\bottomrule
\end{tabular}
}
\caption{
The distribution of cantons of origin in each language subset. No entry means that this language is not spoken in that canton. 
The cantons are ordered in the official order determined by the Swiss Confederation (mostly based on the date of entry into the confederation).
High-resource cantons (> 20\% of decisions per language) are marked in bold. Low-resource cantons (< 5\% of decisions per language) are underlined.
}
\label{tab:origin_canton_distribution}
\vspace{-4mm}
\end{table}

\section{Methods}
\label{sec:methods}

\subsection{Baselines} We first experiment with three baselines. The first one is a \emph{majority} baseline that selects the majority (\emph{dismissal}) class always across cases. The \emph{stratified} baseline predicts labels randomly, respecting the training distribution. The last baseline is a \emph{linear} classifier relying on \ac{TF-IDF} features for the 35K most frequent n-grams in the training set.

\subsection{BERT-based methods} 

BERT \cite{devlin_bert_2019} and its variants \cite{yang_xlnet_2020, liu_roberta_2019, lan_albert_2020}, inter alia, dominate \ac{NLP} as state-of-the-art in many tasks \cite{wang_glue_2018, wang_superglue_2019}. Hence, we examine an arsenal of BERT-based methods.\vspace{2mm}

\noindent\textbf{Standard BERT}
We experimented with monolingual BERT models for German \cite{chan_deepset_2019}, French \cite{martin_camembert_2020} and Italian  \cite{parisi_umberto_2020} and also the multilingual BERT of \cite{devlin_bert_2019}. 
Since the facts are often longer than 512 tokens (see Section \ref{sec:data_description} for details), there is a need to adapt the models to long textual input.\vspace{2mm}

\noindent\textbf{Long BERT}
is an extension of the standard BERT models, where we extend the maximum sequence length by introducing additional positional embeddings. In our case, the additional positional encodings have been initialized by replicating the original pre-trained 512 ones 4 times (2048 in total). While Long BERT can process the full text in the majority of the cases, its extension leads to longer processing time and higher memory requirements.\vspace{2mm}

\noindent\textbf{Hierarchical BERT}, similar to the one presented in \citet{chalkidis_neural_2019}, uses a shared standard BERT encoder processing segments up to 512 tokens to encode each segment independently. To aggregate all (in our case 4) segment encodings, we pass them through an additional \ac{BiLSTM} encoder and concatenate the final LSTM output states to form a single document representation for classification.

\begin{table*}[!ht]
\resizebox{\textwidth}{!}{%
\begin{tabular}{@{}lrrcrrcrr@{}}
\toprule
Model           & \multicolumn{2}{c}{de} & \phantom{a}  & \multicolumn{2}{c}{fr} & \phantom{a}  & \multicolumn{2}{c}{it}\\
\cmidrule{2-3} \cmidrule{5-6} \cmidrule{8-9}
& Micro-F1$\uparrow$ & Macro-F1$\uparrow$ && Micro-F1$\uparrow$ & Macro-F1$\uparrow$ && Micro-F1$\uparrow$ & Macro-F1$\uparrow$\\ 
\midrule
\multicolumn{3}{l}{\emph{baselines}}\\
\midrule
Majority       & \textbf{80.3} & 44.5                  && \textbf{81.5} & 44.9                 && \textbf{81.3} & 44.8\\
Stratified          & 66.7 $\pm$ 0.3 & 50.0 $\pm$ 0.4         && 66.3 $\pm$ 0.2 & 50.0 $\pm$ 0.4        && 69.9 $\pm$ 1.8 & 48.8 $\pm$ 2.4\\
Linear (BoW) & 65.4 $\pm$ 0.2 & 52.6 $\pm$ 0.1 && 71.2 $\pm$ 0.1 & 56.6 $\pm$ 0.2 && 67.4 $\pm$ 0.5 & 53.9 $\pm$ 0.6   \\
\midrule
\multicolumn{3}{l}{\emph{standard} (up to 512 tokens)}\\
\midrule
Native BERT         & 74.0 $\pm$ 4.0 & 63.7 $\pm$ 1.7           && 74.7 $\pm$ 1.8 & 58.6 $\pm$ 0.9      && 76.1 $\pm$ 3.7 & 55.2 $\pm$ 3.7\\
Multilingual BERT   & 68.4 $\pm$ 5.1 & 58.2 $\pm$ 4.8       && 71.3 $\pm$ 4.3 & 55.0 $\pm$ 0.8        && 77.6 $\pm$ 2.4 & 53.0 $\pm$ 1.1\\
\midrule
\multicolumn{3}{l}{\emph{long} (up to 2048 tokens)}\\
\midrule
Native BERT         & 76.5 $\pm$ 3.7 & 67.9 $\pm$ 1.8       && 77.2 $\pm$ 3.4  & 68.0 $\pm$ 1.8       && 77.1 $\pm$ 3.9 & \textbf{59.8 $\pm$ 4.6}\\
Multilingual BERT   & 75.9 $\pm$ 1.6 & 66.5 $\pm$ 0.8       && 73.3 $\pm$ 1.9 & 64.3 $\pm$ 1.5      && 76.0 $\pm$ 2.6 & 58.4 $\pm$ 3.5\\
\midrule
\multicolumn{3}{l}{\emph{hierarchical} (two-tier 4$\times$ 512 tokens)}\\
\midrule
Native BERT         & 77.1 $\pm$ 3.7 & \textbf{68.5 $\pm$ 1.6} && 80.2 $\pm$ 2.0 & \textbf{70.2 $\pm$ 1.1} && 75.8 $\pm$ 3.5 & 57.1 $\pm$ 6.1\\
Multilingual BERT   & 76.8 $\pm$ 3.2 & 57.1 $\pm$ 0.8       && 76.3 $\pm$ 4.1 & 67.2 $\pm$ 2.9      && 72.4 $\pm$ 16.6 & 55.5 $\pm$ 9.5\\
\bottomrule
\end{tabular}
}
\caption{All the models have been trained and evaluated in the same language. With \emph{Native BERT} we mean the BERT model pre-trained in the respective language. The best scores for each language are in bold. Given the high class imbalance, BERT-based methods under-perform in Micro-F1 compared to the \emph{Majority} baseline, while being substantially better in Macro-F1.
}
\label{tab:monolingual_results}
\vspace{-4mm}
\end{table*}

\section{Experiments}
\label{sec:experiments}

In this Section, we describe the conducted experiments alongside the presentation of the results and an analysis of the results of the German dataset in terms of diachronicity (judgment year), legal area, input (text) length and canton of origin.

\subsection{Experimental SetUp}
\label{sec:experimental_setup}

During training, we over-sample the cases representing the minority class (\emph{approval}).\footnote{In preliminary experiments, we find that this sampling methodology outperforms both the standard Empirical Risk Minimization (ERM) and the class-wise weighting of the loss penalty, i.e., considering each class loss 50-50.} 
Across BERT-based methods, we use Early Stopping on development data, an initial learning rate of 3e-5 and batch size 64 across experiments.
The standard BERT models have been trained and evaluated with maximum sequence length 512 and the two variants of BERT with maximum sequence length 2048. 
The 2048 input length has been chosen based on a balance between memory and compute restrictions and the statistics of the length of facts (see Section \ref{sec:long_documents}), where we see that the vast majority of cases contains less than 2K tokens. 
Additionally, this gives us the possibility to investigate differences by input (text) length (see Section \ref{sec:exp_input_length}).
We report both micro- and macro-averaged F1-score on the test set.
Micro-F1 is averaged across samples whereas Macro-F1 is averaged across samples inside each class and then across the classes. 
Therefore, a test example in a minority class has a higher weight in Macro-F1 than an example from the majority class.
In classification problems with imbalanced class distributions (such as the one we examine), Macro-F1 is more realistic than Micro-F1 given that we are equally interested in both classes. 
Each experiment has been run with 5 different random seeds.
We report the average score and standard deviation across experiments.
The experiments have been performed on a single GeForce RTX 3090 GPU with mixed precision and gradient accumulation.
We used the Hugging Face Transformers library \cite{wolf_transformers_2020} and the BERT models available from \url{https://huggingface.co/models}.

\begin{table*}[t]
\resizebox{\textwidth}{!}{%
\begin{tabular}{@{}lrrrrcrrcrr@{}}
\toprule
\multicolumn{3}{c}{Legal Area} & \multicolumn{2}{c}{standard} & \phantom{a}  & \multicolumn{2}{c}{long} & \phantom{a}  & \multicolumn{2}{c}{hierarchical}\\
\cmidrule{4-5} \cmidrule{7-8} \cmidrule{10-11}
Legal Area & \# cases & approval rate & Micro-F1$\uparrow$ & Macro-F1$\uparrow$ && Micro-F1$\uparrow$ & Macro-F1$\uparrow$ && Micro-F1$\uparrow$ & Macro-F1$\uparrow$\\ 
\midrule
\underline{public law} & 2587 & 20.6\% &  66.6 $\pm$ 6.2 & 53.1 $\pm$ 1.8 && 64.6 $\pm$ 6.7 & 53.8 $\pm$ 2.1 && 64.8 $\pm$ 8.1 & 53.7 $\pm$ 3.0\\
\textbf{penal law} & 2900 & 21.0\% &  83.6 $\pm$ 1.8 & 74.8 $\pm$ 1.5 && 87.6 $\pm$ 1.6 & 81.1 $\pm$ 2.3 && 88.4 $\pm$ 1.0 & 82.6 $\pm$ 2.5\\
social law & 661 & 19.3\% &  71.1 $\pm$ 4.3 & 65.2 $\pm$ 2.6 && 74.8 $\pm$ 4.0 & 69.1 $\pm$ 2.8 && 75.4 $\pm$ 3.9 & 69.4 $\pm$ 2.5\\
civil law & 1574 & 16.5\%  &  73.6 $\pm$ 4.8 & 55.5 $\pm$ 1.0 && 79.0 $\pm$ 3.4 & 65.1 $\pm$ 2.4 && 78.9 $\pm$ 3.8 & 65.9 $\pm$ 2.8\\
\bottomrule
\end{tabular}
}
\caption{We used the German native BERT model pre-trained and evaluated on the German data. In the German test set there are no insurance law cases and only 3 cases with other legal areas. The area where models perform best is in bold and the area where they perform worst is underlined.}
\label{tab:legal_areas}
\end{table*}

\subsection{Main Results}

Table \ref{tab:monolingual_results} shows the results across methods for all language subsets. 
We observe that the native BERT models outperform their multi-lingual counterpart; while not being domain-specific, these models can still better model the case facts. 
Given the high class imbalance, all BERT-based methods under-perform in Micro-F1, being biased towards \emph{dismissal} performance compared to the naive Majority baseline, while doing substantially better in Macro-F1. 
Hierarchical and Long BERT-based methods consistently out-perform the linear classifiers across languages (+10\% in Macro-F1), while standard BERT is comparable or better than linear models, although it considers only up to 512 tokens.
While performance of BERT-based methods is quite comparable between the German and French subsets with 35K and 21K training samples respectively, it is far worse in the Italian subset, where there are only 3K training samples. 
In two out of three languages (German and French with 20K+ training samples) hierarchical BERT has borderline better performance compared to long BERT (+1.6-2.2\% in Macro-F1), but in both cases the difference is very close to the error margin (standard deviation). 
We would like to remark that the results of Hierarchical BERT could possibly be improved considering a finer (more intuitive) segmentation of the text into sentences or paragraphs.\footnote{Currently, we segment the text into chunks of 512 tokens to avoid excessive padding that will further increase the needed number of segments and will lead to even higher time and memory demands.} 
We leave the investigation for alternative text segmentation schemes for future work.

\subsection{Discussion - Bivariate Analysis}

In this section, we analyze the results in relation to specific attributes (publication year, input (text) length, legal area and canton of origin) in order to evaluate the model robustness and identify how specific aspects affect the model performance.\vspace{2mm}

\pgfplotstableread{
name mean error
{2017} 64.2 2.1
{2018} 63.3 1.2
{2019} 64.8 1.9
{2020} 62.4 1.6
}\temporalShiftStandard
\pgfplotstableread{
name mean error
{2017} 69.1 2.4
{2018} 67.1 1.8
{2019} 67.5 1.7
{2020} 67.8 1.8
}\temporalShiftLong
\pgfplotstableread{
name mean error
{2017} 69.5 2.6
{2018} 67.6 1.9
{2019} 68.3 1.6
{2020} 68.5 1.5
}\temporalShiftHierarchical

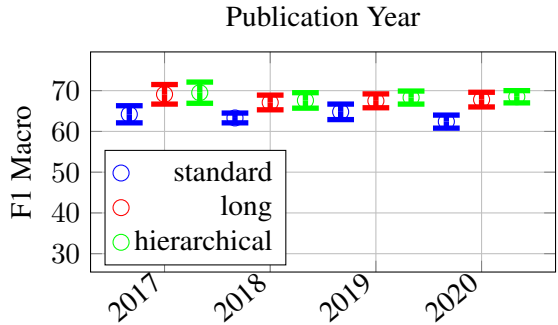
\begin{figure}[t]
\centering
\begin{tikzpicture}
\begin{axis}[
	title={Publication Year},
	legend cell align={right},
	legend pos=south west,
	width=\columnwidth,
	height=4.5cm,
    x tick label style={rotate=40,anchor=east},
    xticklabels from table={\temporalShiftStandard}{name},
    xtick={1, 4, 7, 10, 13},
    enlargelimits=0.10,
    grid=major,
    ylabel={F1 Macro},
    ytick={30, 40, 50, 60, 70},
    ymax=75,
    ymin=30,
]
\addplot 
[color=blue, only marks,mark=o, mark options={scale=1.5}]
plot [
    error bars/.cd,
        y dir=both, y explicit,
        error bar style={line width=2pt},
        error mark options={
            mark size=1pt,
            line width=10pt
        }
    ] table[x expr=\coordindex * 3,y=mean,y error=error]{\temporalShiftStandard};
\addplot 
[color=red, only marks,mark=o, mark options={scale=1.5}]
plot [
    error bars/.cd,
        y dir=both, y explicit,
        error bar style={line width=2pt},
        error mark options={
            mark size=1pt,
            line width=10pt
        }
    ] table[x expr=\coordindex * 3 + 1,y=mean,y error=error]{\temporalShiftLong};
\addplot 
[color=green, only marks,mark=o, mark options={scale=1.5}]
plot [
    error bars/.cd,
        y dir=both, y explicit,
        error bar style={line width=2pt},
        error mark options={
            mark size=1pt,
            line width=10pt
        }
    ] table[x expr=\coordindex * 3 + 2,y=mean,y error=error]{\temporalShiftHierarchical}; 
\legend{
    standard,
    long,
    hierarchical,
}
\end{axis}
\end{tikzpicture}
\caption{This table compares the different BERT types on cases from different years. We used the native German BERT model.}
\label{fig:temporal_shift}
\vspace{-4mm}
\end{figure}

\pgfplotstableread{
name mean error
{1-512} 72.1 1.6
{513-1024} 54.5 2.2
{1025-2048} 50.7 1.0
{2049-4096} 47.3 2.2
{4097-8192} 36.1 2.8
}\inputLengthStandard
\pgfplotstableread{
name mean error
{1-512} 72.2 1.3
{513-1024} 63.4 2.8
{1025-2048} 60.2 2.8
{2049-4096} 50.9 3.6
{4097-8192} 33.1 4.8
}\inputLengthLong
\pgfplotstableread{
name mean error
{1-512} 72.6 1.6
{513-1024} 64.4 2.4
{1025-2048} 61.2 1.3
{2049-4096} 48.0 5.4
{4097-8192} 34.7 5.4
}\inputLengthHierarchical

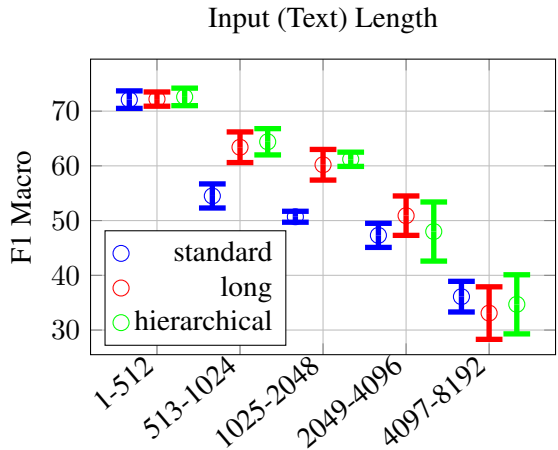
\begin{figure}[!ht]
\centering
\begin{tikzpicture}
\begin{axis}[
	title={Input (Text) Length},
	legend cell align={right},
	legend pos=south west,
	width=\columnwidth,
	height=5.5cm,
    x tick label style={rotate=40,anchor=east},
    xticklabels from table={\inputLengthStandard}{name},
    xtick={1, 4, 7, 10, 13},
    enlargelimits=0.10,
    grid=major,
    ylabel={F1 Macro},
    ytick={30, 40, 50, 60, 70},
    ymax=75,
    ymin=30,
]
\addplot 
[color=blue, only marks,mark=o, mark options={scale=1.5}]
plot [
    error bars/.cd,
        y dir=both, y explicit,
        error bar style={line width=2pt},
        error mark options={
            mark size=1pt,
            line width=10pt
        }
    ] table[x expr=\coordindex * 3,y=mean,y error=error]{\inputLengthStandard};
\addplot 
[color=red, only marks,mark=o, mark options={scale=1.5}]
plot [
    error bars/.cd,
        y dir=both, y explicit,
        error bar style={line width=2pt},
        error mark options={
            mark size=1pt,
            line width=10pt
        }
    ] table[x expr=\coordindex * 3 + 1,y=mean,y error=error]{\inputLengthLong}; 
\addplot 
[color=green, only marks,mark=o, mark options={scale=1.5}]
plot [
    error bars/.cd,
        y dir=both, y explicit,
        error bar style={line width=2pt},
        error mark options={
            mark size=1pt,
            line width=10pt
        }
    ] table[x expr=\coordindex * 3 + 2,y=mean,y error=error]{\inputLengthHierarchical}; 
\legend{
    standard,
    long,
    hierarchical,
}
\end{axis}
\end{tikzpicture}
\caption{This table compares the different long BERT types on different input (text) lengths. We used the native German BERT model.}
\label{fig:input_length}
\vspace{-4mm}
\end{figure}

\subsubsection{Diachronicity}
\label{sec:exp_diachronicity}
In Figure~\ref{fig:temporal_shift}, we present the results grouped by years in the test set (2017-2020). We cannot identify a notable fluctuation in performance across years as there is a very small decrease in performance (approx. -2\% in Macro-F1); most probably because the testing time-frame is really short (4 years). Comparing the performance between the validation (2015-2016) and the test (2017-2020) set (approx. 70\% vs. 68.5\%), again we do not observe an exceptional fluctuation time-wise.\vspace{2mm}

\subsubsection{Input (Text) Length}
\label{sec:exp_input_length}
In Figure~\ref{fig:input_length}, we observe that model performance deteriorates as input (text) length increases, i.e., there is an absolute negative correlation between performance and input (text) length. The two variants of BERT improve results, especially in cases with 512 to 2048 tokens. Since the two variants of BERT have a maximum length of 2048 they perform similar to the standard BERT type in cases longer than 2048 tokens.\vspace{2mm}

\subsubsection{Legal Area}
\label{sec:exp_legal_area}
In Table~\ref{tab:legal_areas}, we observe that the models do not equally perform across legal areas. 
All models seem to be much more accurate in penal law cases, while the performance is much worse (approx. 30\%) in public law cases. 
According to the experts, the jurisprudence in penal law is more united and aligned in Switzerland and outlier judgments are rarer making the task more predictable.
Additionally, in the case of not enough evidence the principle of \emph{``in dubio pro reo''} (reasonable doubt) is applied. \footnote{The principle of \emph{``in dubio pro reo''}, i.e., ``When in doubt, in favor of the defendant.'', is only applicable in penal law cases.}
Another possible reason for the higher performance in penal law could be the increased work performed by the legal clerks in drafting the facts of the case (see Section \ref{sec:facts_vs_considerations}), thus including more useful information relevant to the task.
\vspace{2mm}

\pgfplotstableread{
name mean error
{GE} 59.4 0.9
{VD} 58.8 1.4
{VS} 52.4 2.6
{NE} 57.4 2.9
{FR} 61.1 1.2
{JU} 66.3 2.8
{BE} 48.2 7.7
{CH} 50.0 4.9
}\originCantonStandard
\pgfplotstableread{
name mean error
{GE} 69.4 2.0
{VD} 68.7 1.6
{VS} 63.7 1.2
{NE} 68.0 2.2
{FR} 64.7 3.6
{JU} 69.0 5.1
{BE} 59.9 2.6
{CH} 66.6 7.9
}\originCantonLong
\pgfplotstableread{
name mean error
{GE} 71.8 1.4
{VD} 71.1 1.4
{VS} 64.0 2.6
{NE} 70.8 2.9
{FR} 68.1 2.6
{JU} 74.2 4.5
{BE} 59.2 3.4
{CH} 65.5 5.8
}\originCantonHierarchical

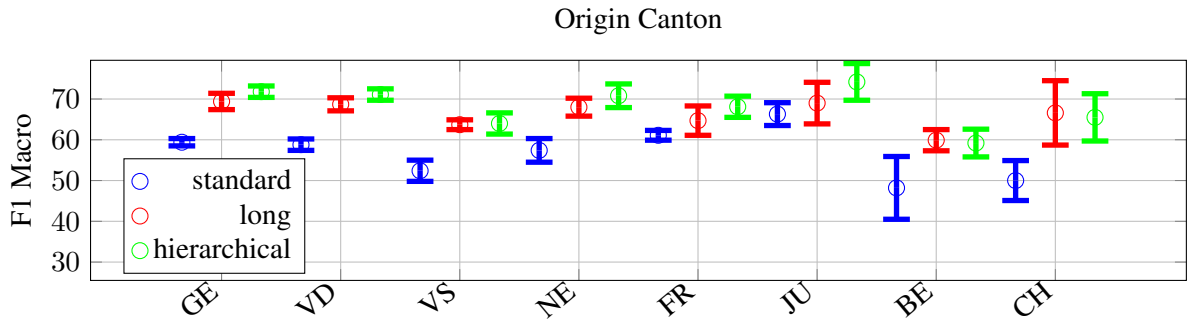
\begin{figure*}[t]
\centering
\begin{tikzpicture}
\begin{axis}[
	title={Origin Canton},
	legend cell align={right},
	legend pos=south west,
	width=\textwidth,
	height=4.5cm,
    x tick label style={rotate=40,anchor=east},
    xticklabels from table={\originCantonStandard}{name},
    xtick={1, 4, 7, 10, 13, 16, 19, 22},
    enlargelimits=0.10,
    grid=major,
    ylabel={F1 Macro},
    ytick={30, 40, 50, 60, 70},
    ymax=75,
    ymin=30,
]
\addplot 
[color=blue, only marks,mark=o, mark options={scale=1.5}]
plot [
    error bars/.cd,
        y dir=both, y explicit,
        error bar style={line width=2pt},
        error mark options={
            mark size=1pt,
            line width=10pt
        }
    ] table[x expr=\coordindex * 3,y=mean,y error=error]{\originCantonStandard};
\addplot 
[color=red, only marks,mark=o, mark options={scale=1.5}]
plot [
    error bars/.cd,
        y dir=both, y explicit,
        error bar style={line width=2pt},
        error mark options={
            mark size=1pt,
            line width=10pt
        }
    ] table[x expr=\coordindex * 3 + 1,y=mean,y error=error]{\originCantonLong}; 
\addplot 
[color=green, only marks,mark=o, mark options={scale=1.5}]
plot [
    error bars/.cd,
        y dir=both, y explicit,
        error bar style={line width=2pt},
        error mark options={
            mark size=1pt,
            line width=10pt
        }
    ] table[x expr=\coordindex * 3 + 2,y=mean,y error=error]{\originCantonHierarchical}; 
\legend{
    standard,
    long,
    hierarchical,
}
\end{axis}
\end{tikzpicture}
\caption{This table compares the different long BERT types on different origin cantons. We used the native French BERT model. The cantons are sorted by the number of cases in the training set descending.}
\label{fig:origin_canton}
\end{figure*}

\subsubsection{Canton of Origin}
\label{sec:exp_canton_of_origin}
In Figure~\ref{fig:origin_canton}, we observe a performance disparity across cantons, although this is neither correlated with the number of cases per canton, nor with the dismissal/approval rate per canton. Thus, the disparity is either purely coincidental and has to do with the difficulty of particular cases in some cantons or there are other factors (e.g., societal, economics) worth considering in future work.\vspace{2mm}

\section{Conclusions \& Future Work}
\label{sec:conclusion}

We introduced a  new multilingual, diachronic dataset of 85K Swiss  Federal  Supreme  Court (\ac{FSCS}) cases, including cases in German, French, and Italian. We presented results considering three alternative BERT-based methods, including methods that can process up to 2048 tokens and thus can read the entirety of the facts in most cases. We found that these methods outperform the standard BERT models and have the best results in Macro-F1, while the naive majority classifier has the best overall results in Micro-F1 due to the high class imbalance of the dataset (more than $\frac{3}{4}$ of the cases are dismissed). Further on, we presented a bivariate analysis between performance and multiple factors (diachronicity, input (text) length, legal area, and canton of origin). The analysis showed that performance deteriorates as input (text) length increases, while the results in cases from different legal areas or cantons vary raising questions on models' robustness under different attributes. 

In future work, we would like to investigate the application of cross-lingual transfer learning techniques, for example the use of Adapters \cite{houlsby2019,pfeiffer2020}. In this case, we could possibly improve the poor performance in the Italian subset, where approx. 3K cases exists, by training a multilingual model across all languages, thus exploiting all available resources, ignoring the traditional language barrier. In the same direction, we could also exploit and transfer knowledge from other annotated datasets that aim at the \ac{LJP} task (e.g., ECtHR and SCOTUS).

More in depth analysis on robustness is also an interesting future avenue. In this direction, we would like to explore distributional robust optimization (DRO) techniques \cite{wilds2021, Wang2021EqualityBT} that aim to mitigate disparities across groups of interest, i.e., labels, cantons and/or legal areas could be both considered in this framework.

Another interesting direction is a deeper analysis with models handling long textual input \cite{beltagy_longformer_2020, zaheer2020bigbird} using alternative attention schemes (window-based, dilated, etc.). Furthermore, none of the examined pre-trained models is legal-oriented, thus pre-training and evaluating such specialized models is also needed, similarly to the English Legal-BERT of \citet{chalkidis-etal-2020-legal}.

\section*{Ethics Statement}

The scope of this work is not to produce a robot lawyer, but rather to study \ac{LJP} in order to broaden the discussion and help practitioners to build assisting technology for legal professionals. We believe that this is an important application field, where research should be conducted \cite{tsarapatsanis-aletras-2021-ethical} to improve legal services and democratize law, while also highlight (inform the audience on) the various multi-aspect shortcomings seeking a responsible and ethical (fair) deployment of technology. In this direction, we provide a well-documented public resource for three languages (German, French, and Italian) that are underrepresented in legal \ac{NLP} literature. We also provide annotations for several attributes (year of publication, legal area, canton/region) and provide a bivariate analysis discussing the shortcomings to further promote new studies in terms of fairness and robustness \cite{Wang2021EqualityBT}, a critical part of \ac{NLP} application in law. All decisions (original material) are publicly available on the entscheidsuche.ch platform and the names of the parties have been redacted (See Figures ~\ref{fig:dismissed_decision} and \ref{fig:approved_decision}) by the court according to its official guidelines\footnote{\url{https://tinyurl.com/mtu23szy} (In German)}.

\section*{Acknowledgements}
This work has been supported by the Swiss National Research Programme “Digital Transformation” (NRP-77)\footnote{\url{https://www.nfp77.ch/en/}} grant number 187477. This work is also partly funded by the Innovation Fund Denmark (IFD)\footnote{\url{https://innovationsfonden.dk/en}} under File No.\ 0175-00011A. We would like to thank: Daniel Kettiger, Magda Chodup, and Thomas Lüthi for their legal advice, Adrian Jörg for help in coding, and Entscheidsuche.ch for providing the data.

\bibliography{main}
\bibliographystyle{acl_natbib}
\clearpage
\appendix

\section{Training Effort}
\label{sec:training_effort}

\begin{table}[!ht]
 \resizebox{\columnwidth}{!}{%
\begin{tabular}{@{}lrrrrr@{}}
\toprule
Type            & BERT          & RoBERTa\\ 
\midrule
standard        & 3.377E+11   & 3.398E+11\\
long            & 1.365E+12   & 1.374E+12\\
hierarchical    & 1.476E+12   & 1.477E+12\\
\bottomrule
\end{tabular}
}
\caption{
This table shows the total floating point operations per epoch per training example used for training each type. Each model has been trained for 2 to 4 epochs (variable because of early stopping). This table can be used to choose a suitable model with limited resources. Additionally, it can be used to measure the environmental impact.
}
\label{tab:training_flops}
\end{table}

Table \ref{tab:training_flops} shows the training effort required for finetuning each type. Training one of the types capable of handling long input results in 4 to 5 times more training operations compared to the standard model. This seems justifiable since the gain from the longer models in terms of F1 score is considerable. Also, the entire cost of finetuning is relatively small.

\begin{figure}[!ht]
\includegraphics[width=\columnwidth]{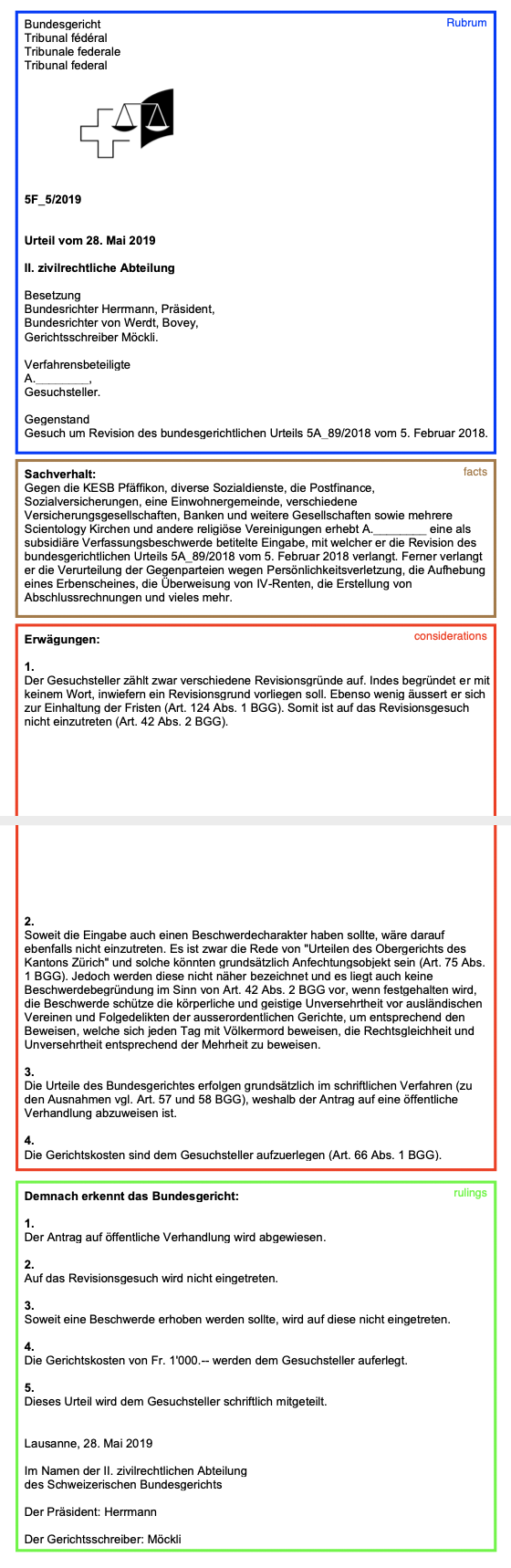}
\caption{This is an example of a dismissed decision: \url{https://tinyurl.com/n44hathc}}
\label{fig:dismissed_decision}
\end{figure}

\begin{figure}[!ht]
\includegraphics[width=\columnwidth]{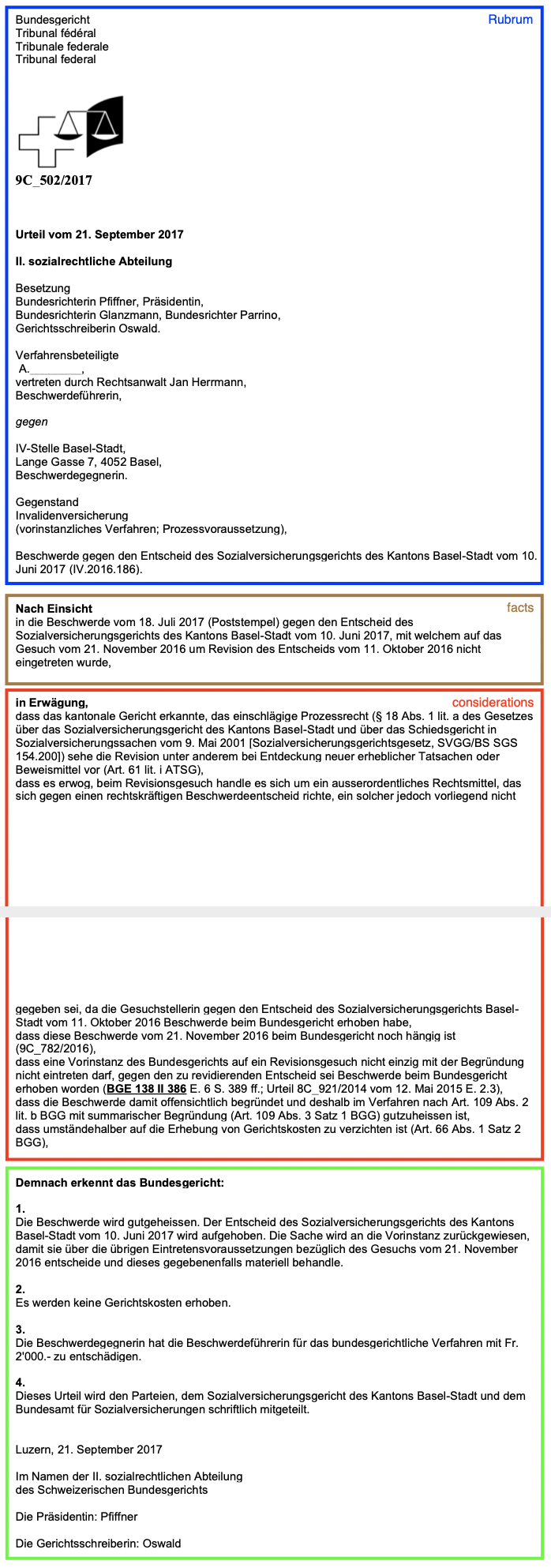}
\caption{This is an example of an approved decision: \url{https://tinyurl.com/mjxfjn65}}
\label{fig:approved_decision}
\end{figure}

\section{Examples}
\label{sec:examples}

In this appendix we show some examples of court decisions with their respective labels.
Figure \ref{fig:dismissed_decision} shows an example of a dismissed decision and Figure \ref{fig:approved_decision} an example of an approved decision. Both decisions are relatively short, but still contain all sections (rubrum, facts, considerations and judgments). They are both very recent, dating from 2019 and 2017 respectively. 

\begin{figure*}
\includegraphics[width=\textwidth]{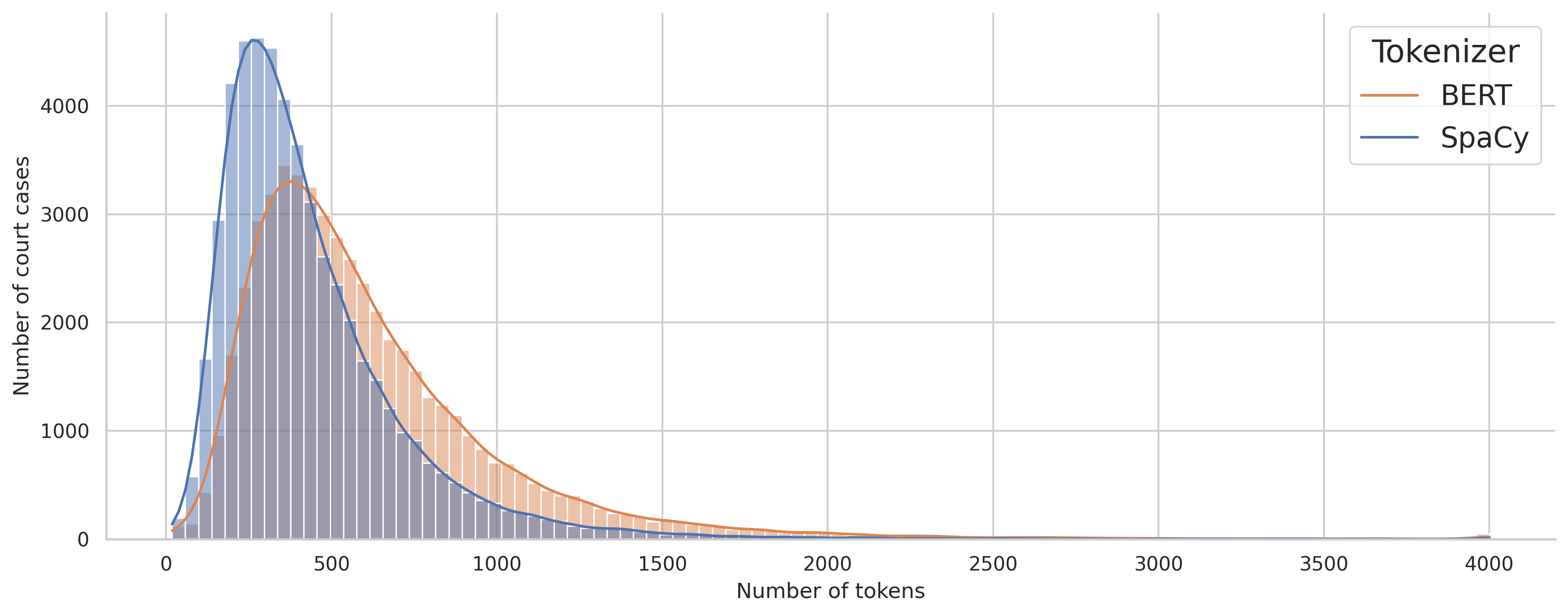}
\caption{This histogram shows the distribution of the input length for German decisions. The blue histogram is generated from tokens generated by the SpaCy tokenizer (regular words). The orange histogram is generated from tokens generated by the SentencePiece tokenizer used in BERT (subword units). Decisions with length over 4000 tokens are grouped in the last bin (before 4000).}
\label{fig:input_length_dist_de}
\end{figure*}

\begin{figure*}
\includegraphics[width=\textwidth]{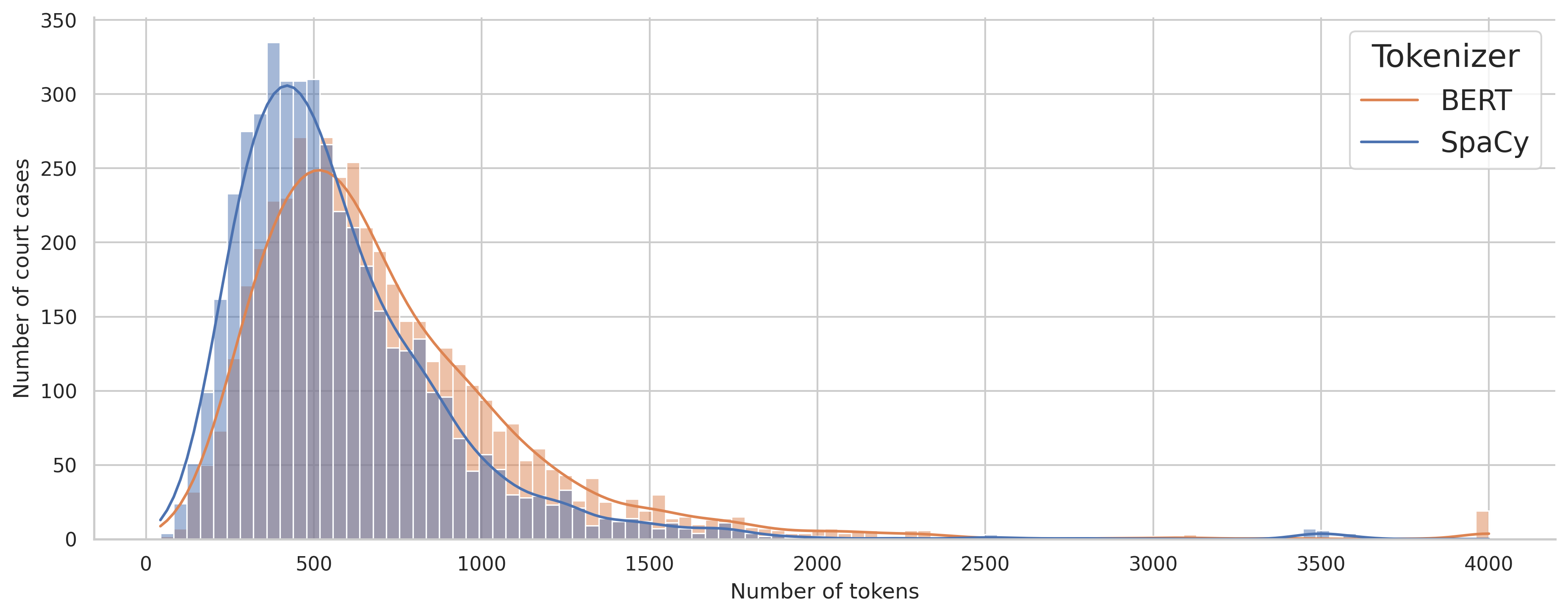}
\caption{This histogram shows the distribution of the input length for Italian decisions. The blue histogram is generated from tokens generated by the SpaCy tokenizer (regular words). The orange histogram is generated from tokens generated by the SentencePiece tokenizer used in BERT (subword units). Decisions with length over 4000 tokens are grouped in the last bin (before 4000).}
\label{fig:input_length_dist_it}
\end{figure*}

\section{Input Length Distribution}
\label{sec:input_length_dist}

In this appendix we show the input length distributions for the German (Figure \ref{fig:input_length_dist_de}) and Italian (Figure \ref{fig:input_length_dist_it}) datasets. We observe that the average Italian decision is longer than the average German decision. Additionally, there is also a higher density in moderately long decisions (over 1000 tokens) and there are many more decisions over 4000 tokens. Apart from the availability of more training data in the German dataset, the shorter decisions may also be an important factor in the better performance we see in most models trained on the German dataset in comparison to the Italian case and to some extent the French case (see Table \ref{tab:monolingual_results}).

\section{Tables to Plots}
\label{sec:tables_to_plots}

In this appendix, we show tables belonging to plots in the main paper to show the exact numbers.
Table \ref{tab:input_length} shows the results regarding the different input lengths. Table \ref{tab:temporal_shift} shows the results regarding different years in the test set. Table \ref{tab:origin_cantons} shows the model performance across different cantons.

\begin{table*}[!ht]
\centering
\resizebox{\textwidth}{!}{%
\begin{tabular}{@{}lrrcrrcrr@{}}
\toprule
Model           & \multicolumn{2}{c}{standard} & \phantom{a}  & \multicolumn{2}{c}{long} & \phantom{a}  & \multicolumn{2}{c}{hierarchical}\\
\cmidrule{2-3} \cmidrule{5-6} \cmidrule{8-9}
& Micro-F1$\uparrow$ & Macro-F1$\uparrow$ && Micro-F1$\uparrow$ & Macro-F1$\uparrow$ && Micro-F1$\uparrow$ & Macro-F1$\uparrow$\\ 
\midrule
1-512 (5479 decisions)    &  81.1 $\pm$ 2.7 & 72.1 $\pm$ 1.6 && 80.8 $\pm$ 2.5 & 72.2 $\pm$ 1.3 && 39.3 $\pm$ 37.2 & 25.1 $\pm$ 17.4\\
513-1024 (3364 decisions) &  65.3 $\pm$ 6.2 & 65.3 $\pm$ 6.2 && 71.8 $\pm$ 5.4 & 63.4 $\pm$ 2.8 && 43.3 $\pm$ 30.8 & 30.5 $\pm$ 13.2\\
1025-2048 (788 decisions) &  63.8 $\pm$ 4.9 & 50.7 $\pm$ 1.0   && 69.1 $\pm$ 5.4 & 60.2 $\pm$ 2.8 && 54.9 $\pm$ 26.7 & 37.2 $\pm$ 15.3\\
2049-4096 (82 decisions) &  64.9 $\pm$ 6.7 & 47.3 $\pm$ 2.2 && 65.1 $\pm$ 9.2 & 50.9 $\pm$ 3.6 && 60.2 $\pm$ 13.3 & 48.0 $\pm$ 5.4\\
4097-8192 (12 decisions) &  56.7 $\pm$ 7.0 & 36.1 $\pm$ 2.8 && 50.0 $\pm$ 10.2 & 33.1 $\pm$ 4.8 && 50.0 $\pm$ 11.8 & 34.7 $\pm$ 5.4\\
\bottomrule
\end{tabular}
}
\caption{Results on the German data grouped by text length. Performance deteriorates as text length is increased.}
\label{tab:input_length}
\end{table*}

\begin{table*}[!ht]
\resizebox{\textwidth}{!}{%
\begin{tabular}{@{}lrrcrrcrr@{}}
\toprule
Model           & \multicolumn{2}{c}{standard} & \phantom{a}  & \multicolumn{2}{c}{long} & \phantom{a}  & \multicolumn{2}{c}{hierarchical}\\
\cmidrule{2-3} \cmidrule{5-6} \cmidrule{8-9}
& Micro-F1$\uparrow$ & Macro-F1$\uparrow$ && Micro-F1$\uparrow$ & Macro-F1$\uparrow$ && Micro-F1$\uparrow$ & Macro-F1$\uparrow$\\ 
\midrule
2017 &  73.9 $\pm$ 4.2 & 64.2 $\pm$ 2.1 && 77.1 $\pm$ 3.9 & 69.1 $\pm$ 2.4 && 77.4 $\pm$ 3.9 & 69.5 $\pm$ 2.6\\
2018 &  74.2 $\pm$ 3.8 & 63.3 $\pm$ 1.2 && 76.6 $\pm$ 3.7 & 67.1 $\pm$ 1.8 && 76.7 $\pm$ 4.0 & 67.6 $\pm$ 1.9\\
2019 &  74.5 $\pm$ 4.0 & 64.8 $\pm$ 1.9 && 76.0 $\pm$ 3.7 & 67.5 $\pm$ 1.7 && 76.9 $\pm$ 3.8 & 68.3 $\pm$ 1.6\\
2020 &  73.5 $\pm$ 4.2 & 62.4 $\pm$ 1.6 && 76.6 $\pm$ 3.4 & 67.8 $\pm$ 1.8 && 77.4 $\pm$ 3.1 & 68.5 $\pm$ 1.5\\
\bottomrule
\end{tabular}
}
\caption{We used the German native BERT model pretrained and evaluated on the German data.}
\label{tab:temporal_shift}
\end{table*}

\begin{table*}[!ht]
\centering
\resizebox{\textwidth}{!}{%
\begin{tabular}{@{}lrrrrcrrcrr@{}}
\toprule
\multicolumn{3}{c}{Canton}           & \multicolumn{2}{c}{standard} & \phantom{a}  & \multicolumn{2}{c}{long} & \phantom{a}  & \multicolumn{2}{c}{hierarchical}\\
\cmidrule{4-5} \cmidrule{7-8} \cmidrule{9-10}
Canton & \# cases & approval rate & Micro-F1$\uparrow$ & Macro-F1$\uparrow$ && Micro-F1$\uparrow$ & Macro-F1$\uparrow$ && Micro-F1$\uparrow$ & Macro-F1$\uparrow$\\ 
\midrule
Berne (BE)                  & 332   & 9.5\%     &  79.4 $\pm$ 4.6 & 48.2 $\pm$ 7.7 && 78.7 $\pm$ 4.7 & 59.9 $\pm$ 2.6 && 78.5 $\pm$ 2.7 & 59.2 $\pm$ 3.4\\
Fribourg (FR)               & 1121  & 14.7\%    &  76.7 $\pm$ 3.1 & 61.1 $\pm$ 1.2 && 75.8 $\pm$ 5.2 & 64.7 $\pm$ 3.6 && 79.5 $\pm$ 3.4 & 68.1 $\pm$ 2.6\\
Vaud (VD)                   & 5684  & 17.0\%    &  76.0 $\pm$ 1.8 & 58.8 $\pm$ 1.4 && 78.9 $\pm$ 3.0 & 68.7 $\pm$ 1.6 && 82.5 $\pm$ 1.7 & 71.1 $\pm$ 1.4\\
Valais (VS)                 & 1399  & 20.6\%    &  75.1 $\pm$ 1.0 & 52.4 $\pm$ 2.6 && 75.0 $\pm$ 2.6 & 63.7 $\pm$ 1.2 && 76.1 $\pm$ 3.3 & 64.0 $\pm$ 2.6\\
Neuchâtel (NE)              & 1226  & 14.9\%    &  76.2 $\pm$ 3.6 & 57.4 $\pm$ 2.9 && 79.0 $\pm$ 3.9 & 68.0 $\pm$ 2.2 && 82.3 $\pm$ 2.7 & 70.8 $\pm$ 2.9\\
Genève (GE)                 & 6017  & 21.8\%    &  72.0 $\pm$ 3.1 & 59.4 $\pm$ 0.9 && 76.0 $\pm$ 3.3 & 69.4 $\pm$ 2.0 && 79.4 $\pm$ 2.3 & 71.8 $\pm$ 1.7\\
Jura (JU)                   & 425   & 15.7\%    &  80.1 $\pm$ 3.2 & 66.3 $\pm$ 2.8 && 78.9 $\pm$ 5.8 & 69.0 $\pm$ 5.1 && 83.8 $\pm$ 4.3 & 74.2 $\pm$ 4.5\\
Swiss Confederation (CH)    & 227   & 26.7\%    &  70.0 $\pm$ 2.7 & 50.0 $\pm$ 4.9 && 72.0 $\pm$ 8.7 & 66.6 $\pm$ 7.9 && 73.3 $\pm$ 4.4 & 65.5 $\pm$ 5.8\\
\bottomrule
\end{tabular}
}
\caption{We used the French native BERT model pretrained and evaluated on the French data. The number of cases is counted on the training set per canton. The approval rate is calculated on the test set.}
\label{tab:origin_cantons}
\end{table*}

\section{Training with Class Weights}
\label{sec:training_with_class_weights}

In this appendix we show the results of training the models with class weights instead of oversampling.
Table \ref{tab:monolingual_results_class_weights} shows the training results. We notice, that for many configurations (especially with XLM-R), the model only learns the majority classifier. This leads to a very low Macro-F1 score. We also experimented with undersampling as an alternative to oversampling, but saw similar results to the training with class weights. 

\begin{table*}[!ht]
\resizebox{\textwidth}{!}{%
\begin{tabular}{@{}lrrcrrcrr@{}}
\toprule
Model           & \multicolumn{2}{c}{de} & \phantom{a}  & \multicolumn{2}{c}{fr} & \phantom{a}  & \multicolumn{2}{c}{it}\\
\cmidrule{2-3} \cmidrule{5-6} \cmidrule{8-9}
& Micro-F1$\uparrow$ & Macro-F1$\uparrow$ && Micro-F1$\uparrow$ & Macro-F1$\uparrow$ && Micro-F1$\uparrow$ & Macro-F1$\uparrow$\\ 
\midrule
\emph{baselines}\\
Most Frequent   & 80.3 & 44.5                         && 81.5 & 44.9                        && 81.3 & 44.8\\
Stratified      & 66.7 $\pm$ 0.3 & 50 $\pm$ 0.4     && 66.3 $\pm$ 0.2 & 50 $\pm$ 0.4        && 69.9 $\pm$ 1.8 & 48.8 $\pm$ 2.4\\
Uniform         & 50 $\pm$ 0.3 & 44.8 $\pm$ 0.4     && 50 $\pm$ 0.6 & 44.5 $\pm$ 0.5        && 49.7 $\pm$ 2.4 & 44 $\pm$ 2.3\\
\emph{standard}\\
Native BERT     & 71.1 $\pm$ 3.3 & 62.6 $\pm$ 1.6    && 72.8 $\pm$ 5.5 & 58.2 $\pm$ 1.2  && 67 $\pm$ 13.1 & 49.4 $\pm$ 5.1\\
XLM-RoBERTa     & 77.8 $\pm$ 6.3 & 47.3 $\pm$ 6.3   && 76.1 $\pm$ 7.4 & 48.4 $\pm$ 4.9   && 80.4 $\pm$ 1.9 & 44.7 $\pm$ 0.4\\
\emph{long}\\
Native BERT     & 81.9 $\pm$ 1.2 & \textbf{69.5 $\pm$ 0.9} && \textbf{81.8 $\pm$ 1.5} & 69.4 $\pm$ 1.7 && 80.2 $\pm$ 1.4 & 46.1 $\pm$ 2.2\\
XLM-RoBERTa     & \textbf{81.5 $\pm$ 0.7} & 59.4 $\pm$ 9.6             && 81.5 $\pm$ 0.5 & 51.3 $\pm$ 8.8 && 81.3 & 44.8\\
\emph{hierarchical}\\
Native BERT     & 78.6 $\pm$ 2.1 & 69.2 $\pm$ 0.6    && 79.3 $\pm$ 0.8 & \textbf{70 $\pm$ 0.7} && 80.6 $\pm$ 1.1 & \textbf{50.5 $\pm$ 6.5}\\
XLM-RoBERTa     & 80.3 & 44.5                          && 80.3 $\pm$ 1.8 & 49.6 $\pm$ 9.8 && 81.3 & 44.8\\
\bottomrule
\end{tabular}
}
\caption{All the models have been trained and evaluated in the same language. With \emph{Native BERT} we mean the BERT model pretrained in the respective language. The \emph{Most Frequent} baseline just selects the majority class always. The \emph{Stratified} baseline predicts randomly, respecting the training distribution. The best scores for each language are in bold. To combat label imbalance, we weighted the minority class samples more in the loss function.
}
\label{tab:monolingual_results_class_weights}
\end{table*}

\section{Classifier Confidence}
\label{sec:classifier_confidence}

In this appendix, we discuss the reliability of the confidence scores of the classifier output alongside the predictions.
The confidence scores are computed by taking the softmax on the classifier outputs, so that we get a probability (confidence) score of a given class between 0 and 100.
The hierarchical and long BERT types show an increase in both the confidence in the correct predictions and the incorrect predictions compared to the standard BERT type (with the increase in the correct predictions being more pronounced). This finding holds across all three languages. 

\begin{table*}[!ht]
 \resizebox{\textwidth}{!}{%
\begin{tabular}{@{}lrrcrrcrr@{}}
\toprule
Model           & \multicolumn{2}{c}{de} & \phantom{a}  & \multicolumn{2}{c}{fr} & \phantom{a}  & \multicolumn{2}{c}{it}\\
\cmidrule{2-3} \cmidrule{5-6} \cmidrule{8-9}
& Correct$\uparrow$ & Incorrect$\downarrow$ && Correct$\uparrow$ & Incorrect$\downarrow$ && Correct$\uparrow$ & Incorrect$\downarrow$\\ 
\midrule
standard    & 75.8 $\pm$ 13.6 & 64.7 $\pm$ 10.6 && 71.9 $\pm$ 12.2 & 64.4 $\pm$ 9.8 && 77.6 $\pm$ 12.2 & 68.3 $\pm$ 11.3\\
long        & 78.9 $\pm$ 12.2 & 65.8 $\pm$ 10.9 && 78.3 $\pm$ 11.6 & 67.8 $\pm$ 11.0 && 81.2 $\pm$ 11.2 & 68.4 $\pm$ 10.5\\
hierarchical& 86.6 $\pm$ 15.9 & 69.3 $\pm$ 13.6 && 85.9 $\pm$ 15.2 & 70.8 $\pm$ 13.9 && 88.7 $\pm$ 14.7 & 71.4 $\pm$ 13.4\\
\bottomrule
\end{tabular}
}
\caption{
This table shows the average confidence scores (0-100) of the different types of multilingual BERT models on the test set for correct and incorrect predictions respectively. Both the mean and standard deviation are averaged over 5 random seeds. The model has been finetuned on the entire dataset (all languages) and evaluated on the respective language.
}
\label{tab:classifier_confidence}
\end{table*}

\end{document}